\documentclass[11pt]{article}

\usepackage[final]{acl}

\usepackage{times}
\usepackage{latexsym}
\usepackage[T1]{fontenc}
\usepackage[utf8]{inputenc}
\usepackage{microtype}
\usepackage{inconsolata}

\usepackage{amsmath}
\usepackage{amssymb}
\usepackage{amsfonts}

\usepackage{algorithm}
\usepackage{algorithmic}

\usepackage{graphicx}
\usepackage[table,xcdraw]{xcolor} 
\usepackage{booktabs} 
\usepackage{multirow} 
\definecolor{headerblue}{RGB}{230, 240, 255} 
\definecolor{groupgray}{RGB}{245, 245, 245}  
\definecolor{highlightgreen}{RGB}{235, 255, 235}

\usepackage{tcolorbox}
\definecolor{mainteal}{RGB}{80, 20, 90}      
\definecolor{backteal}{RGB}{250, 245, 252}   

\newtcolorbox{paperbox}[1]{
  colback=backteal,
  colframe=mainteal,
  coltitle=white,          
  boxrule=0.8pt,
  arc=2pt,
  left=4pt, right=4pt, 
  top=3pt, bottom=3pt,     
  fonttitle=\bfseries\small,
  title={#1},
  toptitle=2pt, 
  bottomtitle=0pt,
  before skip=3pt,         
  after skip=0pt           
}

\title{ETR: Outcome-Guided Elastic Trust Regions for Policy Optimization}

\author{
    \textbf{Shijie Zhang}\textsuperscript{1,2,$\dagger$} ,\;
    \textbf{Kevin Zhang}\textsuperscript{2,$\dagger$} ,\;
    \textbf{Zheyuan Gu}\textsuperscript{2,$\dagger$} ,\;
    \textbf{Xiang Guo}\textsuperscript{1} ,\;
    \textbf{Rujun Guo}\textsuperscript{1} ,\;\\
    \textbf{Shaoyu Liu}\textsuperscript{1} ,\;
    \textbf{Guanjun Jiang}\textsuperscript{1} ,\;
    \textbf{Xiaozhao Wang}\textsuperscript{1} \\
    \vspace{0.2cm}
    \textsuperscript{1}Qwen Applications Business Group, Alibaba Group \quad
    \textsuperscript{2}Peking University
}

\begin{document}
\maketitle

\renewcommand{\footnoterule}{}

\renewcommand{\thefootnote}{} 
\footnotetext{\noindent\hspace*{-1.1em} 
\begin{tabular}{@{}l} 
Email: \texttt{\{yaya, kevinzyz\}@stu.pku.edu.cn} \\
\textsuperscript{$\dagger$} denotes core contributors.
\end{tabular}}
\renewcommand{\thefootnote}{\arabic{footnote}} 


\begin{abstract}
Reinforcement Learning with Verifiable Rewards (RLVR) has emerged as an important paradigm for unlocking reasoning capabilities in large language models, exemplified by the success of OpenAI o1 and DeepSeek-R1. Currently, Group Relative Policy Optimization (GRPO) stands as the dominant algorithm in this domain due to its stable training and critic-free efficiency. However, we argue that GRPO suffers from a structural limitation: it imposes a uniform, static trust region constraint across all samples. This design implicitly assumes signal homogeneity, a premise misaligned with the heterogeneous nature of outcome-driven learning, where advantage magnitudes and variances fluctuate significantly. Consequently, static constraints fail to fully exploit high-quality signals while insufficiently suppressing noise, often precipitating rapid entropy collapse. To address this, we propose \textbf{E}lastic \textbf{T}rust \textbf{R}egions (\textbf{ETR}), a dynamic mechanism that aligns optimization constraints with signal quality. ETR constructs a signal-aware landscape through dual-level elasticity: at the micro level, it scales clipping boundaries based on advantage magnitude to accelerate learning from high-confidence paths; at the macro level, it leverages group variance to implicitly allocate larger update budgets to tasks in the optimal learning zone. Extensive experiments on AIME and MATH benchmarks demonstrate that ETR consistently outperforms GRPO, achieving superior accuracy while effectively mitigating policy entropy degradation to ensure sustained exploration.
\end{abstract}



\begin{figure}[h]
    \centering
    \includegraphics[width=0.95\linewidth]{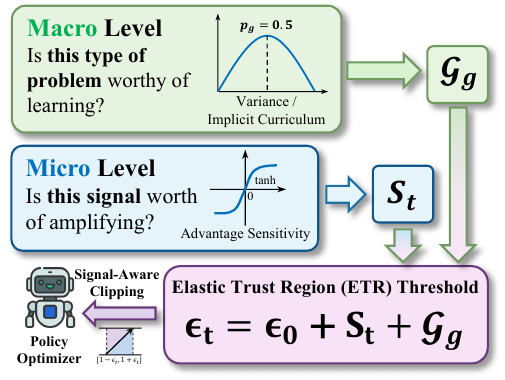}
    \caption{\textbf{Not all learning signals deserve equal trust.} 
    ETR replaces GRPO’s fixed threshold with dynamic bounds that expand at the macro-level for valuable samples and contract at the micro-level for uncertain steps.
    }
    \label{fig:teaser}
\end{figure}

\section{Introduction}

Reinforcement Learning (RL) has become a central technique for improving the reasoning capabilities of Large Language Models (LLMs)~\cite{achiam2023gpt,touvron2023llama, grattafiori2024llama3herdmodels,gemini2.5}, particularly in domains that require multi-step inference, such as mathematics, coding, and logical problem solving~\cite{ouyang2022training,guo2025deepseek,jaech2024openai,shao2024deepseekmath,zhang2025clpocurriculumlearningmeets}. 

Early approaches primarily relied on Reinforcement Learning from Human Feedback (RLHF)~\cite{schulman2017proximal}, where a learned reward model approximates human preferences and guides policy optimization. While effective, reward-model-based pipelines introduce additional sources of noise and biases, and often struggle to provide fine-grained supervision for tasks with objectively verifiable outcomes~\cite{wen2025reinforcementlearningverifiablerewards,wang2025reinforcement}.
To address these limitations, Reinforcement Learning with Verifiable Rewards (RLVR) has recently gained prominence~\cite{wen2025reinforcement,zhang2025markovianreflectiveexplorationbayesadaptive} and is widely applied in existing reasoning models~\cite{guo2025deepseek,kimiteam2025kimik15scalingreinforcement}. 
In RLVR settings, rewards are derived directly from ground-truth outcomes, such as solution correctness or test case pass rates, providing precise and unambiguous optimization signals. This paradigm has proven particularly effective for mathematical reasoning and program synthesis, where correctness can be automatically verified. A popular approach, Group Relative Policy Optimization (GRPO)~\cite{shao2024deepseekmath}, has emerged due to its simplicity and efficiency. By computing relative advantages within sampled groups and eliminating the need for an explicit value function, GRPO significantly reduces computational overhead while retaining strong empirical performance.

Despite its empirical success, GRPO largely inherits from Proximal Policy Optimization’s (PPO)~\cite{schulman2017proximal} static trust region design, enforcing a fixed update constraint across all samples and groups. We argue that this assumption is poorly suited for RLVR, where training signals are inherently heterogeneous. Specifically, advantage values can vary widely in magnitude and sign, and query prompts differ substantially in difficulty, leading to groups with distinct statistical properties. Statistical principles suggest that groups with pass rates near 50\% exhibit maximum variance and contain the richest gradient information\cite{zhang2025clpocurriculumlearningmeets}.
A uniform clipping threshold thus misallocates optimization capacity, over-constraining informative samples while insufficiently regulating low-quality and noisy ones. In practice, this mismatch leads to inefficient learning and rapid policy entropy collapse, thereby limiting exploration and generalization in reasoning tasks.

To address these issues, we \textbf{E}lastic \textbf{T}rust \textbf{R}egions (\textbf{ETR}), a dynamic constraint mechanism that adapts policy updates to outcome statistics. Instead of enforcing a fixed clipping threshold, ETR adjusts the effective trust region based on both sample-level advantages and group-level variance. This design allocates larger update budgets to more informative samples and groups, enabling difficulty-aware optimization and implicitly encouraging a curriculum over the given prompts, applying adequate constraints to groups with varying information density.
ETR integrates seamlessly with GRPO-style objectives and improves training stability without introducing additional networks or supervision.

Our contributions are summarized as follows:
\begin{enumerate}
\item We identify \textit{The Static Mismatch}, a fundamental limitation of static trust regions in GRPO due to heterogeneous, outcome-driven training signals and systematically analyze their impact on policy diversity and learning efficiency.

\item We propose Elastic Trust Regions (ETR), a general dynamic constraint mechanism that adapts policy updates with both sample-level advantages and group-level variance, enabling an implicit curriculum learning over prompt difficulty.

\item Through extensive experiments on mathematical reasoning benchmarks, we demonstrate that ETR consistently outperforms GRPO and its Clip-High variant across all base models. Specifically, on the Qwen3-8B baseline, ETR achieves a \textbf{+7.7\%} gain on AMC23 (\textbf{+5.6\%} over Clip-High) and a \textbf{+2.6\%} gain on AIME25 (\textbf{+2.3\%} over Clip-High) over GRPO, while also showing robust generalization on out-of-distribution tasks and substantially mitigating rapid policy entropy decay.
\end{enumerate}

\begin{figure}[h]
    \centering
    \includegraphics[width=\linewidth]{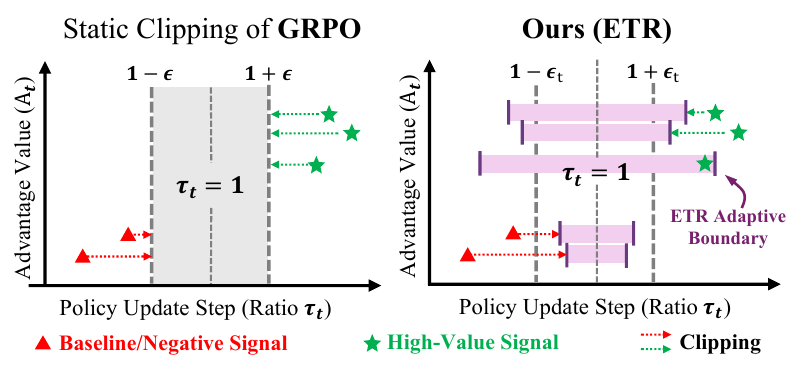}
    \caption{Comparison of optimization landscapes between GRPO and ETR.
    \textbf{Left (GRPO):} the static clipping range suffers from \textit{static mismatch}, where high-value signals (green stars, large advantage) are aggressively truncated, limiting efficient learning. \textbf{Right (ETR):} our method creates a signal-aware adaptive boundary. By scaling the trust region proportional to signal strength, ETR successfully encapsulates both high-advantage samples while constraining negative signals (red triangles) to ensure stability.}
    \label{fig:placeholder}
\end{figure}
\section{Preliminaries}

We consider the Reinforcement Learning with Verifiable Rewards (RLVR) setting for LLMs. The policy $\pi_\theta$ generates a response $o$ given a query $q$.

\paragraph{Proximal Policy Optimization (PPO).}
PPO~\cite{schulman2017proximal} stabilizes training by constraining policy updates within a trust region. It employs a clipped surrogate objective to prevent destructive large updates. The objective function is defined as:
\begin{equation}
    \mathcal{J}_{PPO}(\theta) = \mathbb{E}_t \left[ \min \left( r_t(\theta) A_t, \ L_t(\theta) A_t \right) \right]
\end{equation}
\begin{equation}
    L_t(\theta) = \text{clip}(r_t(\theta), 1-\epsilon, 1+\epsilon)
\end{equation}
where $r_t(\theta) = \frac{\pi_\theta(a_t|s_t)}{\pi_{\theta_{old}}(a_t|s_t)}$ is the probability ratio. To estimate the advantage $A_t$, PPO typically relies on a learned value function $V_\phi(s)$ and utilizes Generalized Advantage Estimation (GAE)~\cite{schulman2015high} to balance bias and variance.

\paragraph{Group Relative Policy Optimization (GRPO).}
GRPO~\cite{shao2024deepseekmath} eliminates the need for a value network. For each query $q$, it samples a group of $G$ outputs $\{o_1, \dots, o_G\}$ from the old policy $\pi_{\theta_{old}}$. The advantage $A_i$ is computed by normalizing the intra-group rewards in $G$.

The objective function of GRPO can be divided into two main parts: i) a clipped surrogate objective (to limit the update step from $\pi_{\theta_{old}}$) (that maximizes the reward / advantage of the sampled group, while the clipping limits the updated gradient step) and ii) a KL penalty term (to keep the policy close to the reference model $\pi_{ref}$).
\begin{equation}
\begin{split}
    \mathcal{J}_{GRPO}(\theta) = \mathbb{E} \Bigg[ \frac{1}{G} \sum_{i=1}^G \frac{1}{|o_i|} \sum_{t=1}^{|o_i|} \bigg( \min \Big( r_t A_t, \\
    \text{clip}(r_t, 1-\epsilon, 1+\epsilon) A_t \Big) - \beta D_{KL}(\pi_\theta || \pi_{ref}) \bigg) \Bigg]
\end{split}
\end{equation}
where $r_t = \frac{\pi_\theta(o_t|q, o_{<t})}{\pi_{\theta_{old}}(o_t|q, o_{<t})}$ is the probability ratio between the current and old policies, and $\beta$ is the coefficient for the KL regularization term towards the reference model.
\section{Methodology}

In this section, we formalize the optimization landscape of reasoning tasks. We first identify the theoretical limitations of static clipping (Sec \ref{meth:theory}) and derive the optimal dynamic boundary from a weighted trust region framework (Sec \ref{meth:tr}). Finally, we instantiate this framework into the Elastic Trust Regions (ETR) algorithm (Sec \ref{meth:etr}).

\begin{figure*}[t]
    \centering
    \includegraphics[width=0.95\linewidth]{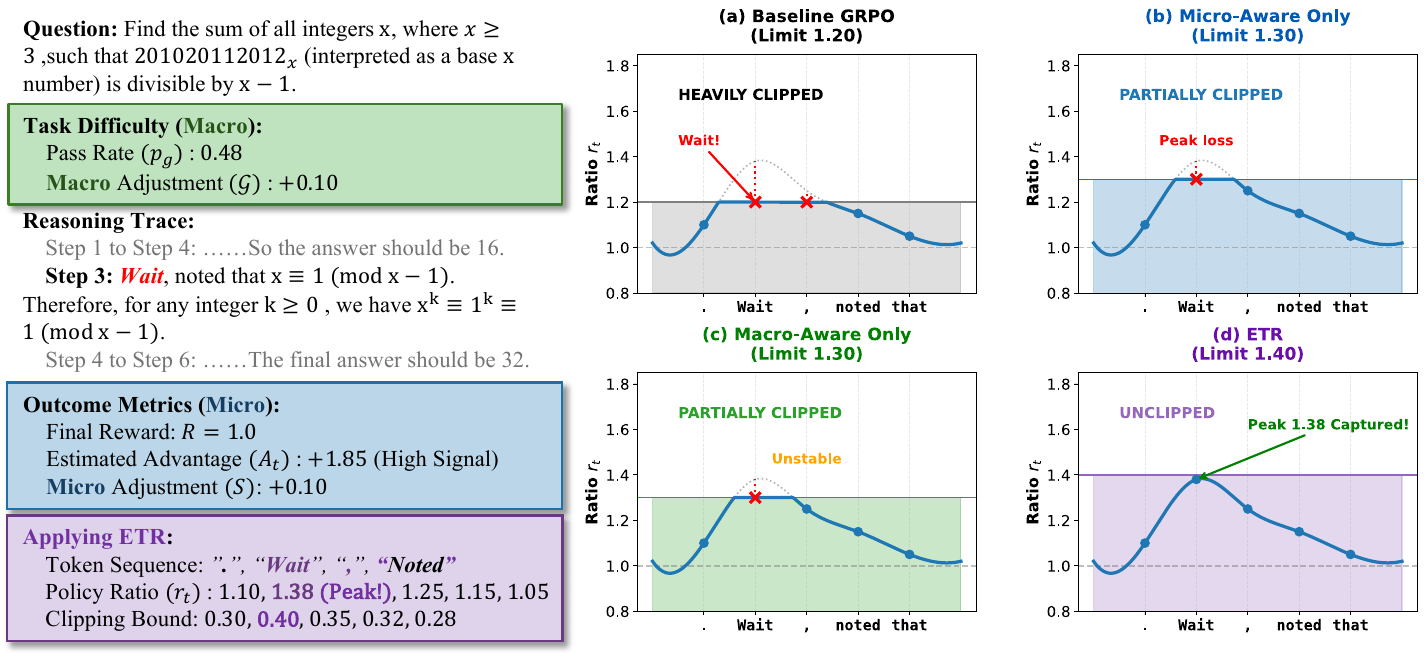}
    \caption{\textbf{A case study on a high-difficulty reasoning step.} 
    We visualize the policy ratio trajectory for a critical step that requires a large update ($\approx 1.38$).
    \textbf{(a)} GRPO (limit 1.20) clips the gradient greatly, losing information.
    \textbf{(b) \& (c)} Applying only Micro or Macro adjustments alleviates this issue, but still results in some clipping.
    \textbf{(d) Full ETR} combines both adjustments, raising the dynamic boundary to 1.40 to capture the peak learning signal. This demonstrates how ETR prevents the loss of crucial information in important sparse-reward reasoning tasks.}
    \label{fig:method_overview}
\end{figure*}

\subsection{Theoretical Diagnosis: The Static Mismatch}
\label{meth:theory}

To understand the limitations of current RLVR methods, we revisit the local optimization objective of Trust Region methods. For a single sample, the objective with a KL divergence penalty $\beta$ is approximated via a second-order Taylor expansion as $\max_{r} r \cdot A - \frac{\beta}{2}(r-1)^2$. Solving for the optimal update step $r^*$, we obtain a linear relationship $|r^* - 1| \propto |A|/\beta$.

This relationship reveals that the optimal update magnitude is inherently dependent on the signal strength $A$. 
Crucially, this implies that the ideal trust region is not a constant, but a dynamic variable determined by the confidence of the policy's evaluation.
However, existing algorithms like PPO and GRPO enforce a uniform, fixed constraint $|r-1| \le \epsilon$, disregarding the specific utility of each token.
This imposes a "one-size-fits-all" limitation: updates driven by high-quality reasoning paths are inadequately capped, while updates from ambiguous or noisy samples are allowed disproportionate freedom.
We formalize this structural inefficiency as the \textit{Static Mismatch}:

\begin{paperbox}{Definition 3.1: The Static Mismatch}
The Static Mismatch denotes the misalignment between a fixed threshold $\epsilon$ and the heterogeneous nature of training signals. \textbf{Micro-level (Magnitude):} For high-advantage samples, the theoretical optimal step exceeds the static bound ($|r^*-1| > \epsilon$), leading to aggressive gradient truncation and the under-utilization of golden signals. \textbf{Macro-level (Variance):} The fixed threshold ignores the varying difficulty across problem groups, applying identical constraints to distributions with vastly different variance profiles and information densities.
\end{paperbox}
\subsection{Theoretical Framework: Signal-Aware Trust Regions}
\label{meth:tr}

To resolve this mismatch, we propose to replace the uniform trust region assumption with a \textit{Signal-Aware Weighted Constraint}. Instead of enforcing a uniform budget $\mathbb{E}[D_{KL}] \le \delta$ for all data points, we introduce a scaling factor $\rho_t \ge 1$ representing the signal quality of the $t$-th token:
\begin{equation}
    \mathbb{E}_{t} \left[ \frac{1}{\rho_t} D_{KL}(\pi_{\theta_{old}} || \pi_{\theta}) \right] \le \delta
\end{equation}
In this formulation, a larger $\rho_t$ reduces the effective penalty weight of the KL divergence. This justifies allowing a larger deviation for samples with high signal quality. By solving the Lagrangian dual of this constrained optimization problem, we derive the optimal form of the clipping boundary. This theoretical result serves as the cornerstone of our method, explicitly dictating that the trust region radius should not be static, but must scale dynamically with the signal strength to balance stability and efficiency.

\begin{paperbox}{Theorem 3.1: Optimal Dynamic Boundary}
To satisfy the signal-aware weighted constraint, the clipping threshold $\epsilon^*(t)$ for the $t$-th sample must scale with the square root of its signal strength $\rho_t$:
\begin{equation}
    \epsilon_{dynamic}(t) = \epsilon_{base} \cdot \sqrt{\rho_t}
\end{equation}
This theorem provides the mathematical justification for dynamic clipping: the trust region radius should inherently scale with the confidence of the learning signal.
\end{paperbox}

\subsection{Elastic Trust Regions (ETR)}
\label{meth:etr}

Guided by Theorem 3.1, we propose ETR. We parameterize the scaling factor $\rho_t$ using two additive terms to ensure numerical stability and decouple the hyperparameters.

\begin{paperbox}{Algorithm Formulation: Dynamic Threshold $\epsilon_t$}
We define the dynamic clipping threshold $\epsilon_t$ as the base threshold adjusted by micro-level and macro-level elasticity:
\begin{equation}
    \epsilon_t = \epsilon_{base} + \underbrace{\mathcal{S}(A_t)}_{\text{Micro}} + \underbrace{\mathcal{G}(p_g)}_{\text{Macro}}
    \label{eq:etr_epsilon}
\end{equation}
where $\mathcal{S}(A_t)$ accounts for sample-level signal strength, and $\mathcal{G}(p_g)$ accounts for group-level learnability.
\end{paperbox}

\paragraph{Micro-Level Adjustment ($\mathcal{S}$).}
We define the sample-level term to adapt to the magnitude and sign of the advantage. We employ the hyperbolic tangent function to bound the adjustment range.
\begin{itemize}
    \item For positive samples ($A_t > 0$), we relax the upper bound to allow for larger updates on correct paths.
    \item For negative samples ($A_t < 0$), we adjust the lower bound to maintain stability against errors.
\end{itemize}
\begin{equation}
    \mathcal{S}(A_t) = \lambda_1 \cdot \tanh(A_t)
\end{equation}

\paragraph{Macro-Level Adjustment ($\mathcal{G}$).}
We define the group-level term based on the variance of the group outcomes. For a group with pass rate $p_g$, the variance is proportional to $p_g(1-p_g)$.
\begin{equation}
    \mathcal{G}(p_g) = \lambda_2 \cdot 4 p_g (1 - p_g)
\end{equation}

\begin{paperbox}{Insight: Implicit Curriculum Learning}
The macro-level term $\mathcal{G}(p_g)$ reaches its maximum when $p_g \approx 0.5$ (maximum variance) and approaches zero when $p_g \approx 0$ or $1$. This introduces an \textbf{Implicit Curriculum Learning} mechanism: the algorithm automatically allocates a larger update budget to tasks within the ``optimal learning zone'' (high variance), while maintaining conservative constraints on tasks that are currently too hard or too easy.
\end{paperbox}

\paragraph{Final Objective.}
Integrating the dynamic boundary into the GRPO framework, the ETR loss function is computed by substituting $\epsilon_t$ into the standard clipped objective:
\begin{equation}
\begin{aligned}
    \mathcal{J}_{ETR}(\theta) = \mathbb{E} \Bigg[ \frac{1}{G} \sum_{i=1}^G \frac{1}{|o_i|} \sum_{t=1}^{|o_i|} \Big( \min \Big( r_{i,t} A_i, \\
    \text{clip}(r_{i,t}, 1-\epsilon_{i,t}, 1+\epsilon_{i,t}) A_i \Big) - \beta D_{KL}(\pi_\theta || \pi_{ref}) \Big) \Bigg]
\end{aligned}
\end{equation}
Through this mechanism, ETR dynamically aligns the optimization constraints with the statistical properties of the data.
\section{Experiments}

In this section, we empirically evaluate the proposed ETR method on multiple mathematical reasoning benchmarks. Our experiments aim to verify the performance of the dynamic trust region mechanism across different task difficulties, analyze its impact on training convergence and policy exploration, and validate the effectiveness of its individual design components.

\subsection{Experimental Setup}

\paragraph{Benchmarks.}
We assess the reasoning capabilities of the models using a diverse set of datasets. For in-distribution evaluation, we utilize \textbf{MATH500}~\cite{hendrycks2021measuring} and \textbf{AMC23}\cite{li2024numinamath} to represent standard competition-level problems, and \textbf{AIME 2024}\cite{lewkowycz2022solving} and \textbf{AIME 2025} to represent high-difficulty olympiad problems. The AIME datasets contain sparse correct paths, serving as a critical testbed for our signal-aware hypothesis. For these tasks, we report both \textbf{Mean@32} and \textbf{Best@32} accuracy to measure average performance and exploration ceilings, respectively. To evaluate generalization capabilities, we employ out-of-distribution (OOD) benchmarks including \textbf{GPQA-Main}, \textbf{GPQA-Diamond}~\cite{rein2024gpqa}, and \textbf{ACPBench}~\cite{kokel2024acp}, reporting \textbf{Pass@1} accuracy. The training data is derived from the DAPO-Math-17k~\cite{yu2025dapo} dataset.

\paragraph{Baselines and Implementation.}
We conduct experiments using the \texttt{verl} reinforcement learning framework, utilizing FSDP for training and vLLM for inference on 8$\times$H20 GPUs. The experiments cover multiple model architectures: \textbf{Qwen3-8B-Base}~\cite{yang2025qwen3}, \textbf{Llama-3.1-8B-Instruct}~\cite{grattafiori2024llama3herdmodels}, and \textbf{Qwen2.5-7B-Math}~\cite{yang2024qwen25mathtechnicalreportmathematical}.
Our primary baseline is \textbf{Group Relative Policy Optimization (GRPO)} with a static clipping threshold ($\epsilon=0.2$). To investigate the effect of simply relaxing the constraint, we also compare against a \textbf{GRPO-Clip-High} variant inspired by ~\cite{yu2025dapo} with $\epsilon=0.28$.
ETR is implemented within the same codebase, modifying only the clipping logic in the loss function. We set the group size $G=8$ and the sampling temperature to 1.0. The reward function is strictly outcome-based (+1 for correct, -1 for incorrect). All models are trained using the AdamW optimizer. For ETR, the micro-level coefficient $\lambda_1$ and macro-level coefficient $\lambda_2$ are both set to default values of 0.1.

\subsection{Main Results}

Table~\ref{tab:main_results} presents the performance comparison across all benchmarks. ETR consistently outperforms the static GRPO and the Clip-high baselines across different model families. 

\paragraph{Performance on Hard Tasks.}
Notably, the performance gain provided by ETR increases with task difficulty. On the challenging AIME 2024 and AIME 2025 datasets, ETR achieves substantial improvements in the Best@32 metric. This confirms that in hard tasks, where correct reasoning paths are rare and yield high advantage values, the static clipping mechanism leads to under-fitting. ETR's elastic boundary effectively captures these sparse, high-value signals, allowing the model to capitalize on successful exploration.

\paragraph{Out-of-Distribution Generalization.}
ETR also demonstrates consistent improvements on OOD tasks (GPQA and ACPBench). We attribute this generalization capability to the suppression of overfitting. Static clipping constraints can force the model to overfit specific templates in the training set to maximize rewards within a restricted policy space. By tightening constraints on low-signal samples and relaxing them for high-confidence updates, ETR maintains higher policy diversity, encouraging the learning of robust reasoning logic rather than dataset-specific patterns.

\begin{table*}[t]
\centering
\small
\renewcommand{\arraystretch}{1.3}
\setlength{\tabcolsep}{3pt}
\caption{Main results on in-distribution and out-of-distribution benchmarks. In-distribution metrics are reported as \textbf{Mean@32 / Best@32}. \colorbox{highlightgreen}{Highlighted rows} indicate our ETR method. \textbf{Bold} denotes the best performance within each model family.}
\label{tab:main_results}
\begin{tabular}{lccccc|ccc}
\toprule
\multirow{3}{*}{\textbf{Method}} & \multicolumn{5}{c}{\textbf{In-Distribution (Math Reasoning)}} & \multicolumn{3}{c}{\textbf{Out-of-Distribution}} \\ 
 & \multicolumn{5}{c}{\textit{(Mean@32 / Best@32)}} & \multicolumn{3}{c}{\textit{(Pass@1)}} \\
\cmidrule(lr){2-6} \cmidrule(lr){7-9}
 & \textbf{AMC23} & \textbf{AIME24} & \textbf{AIME25} & \textbf{MATH500} & \cellcolor{headerblue}\textbf{Avg.} & \textbf{GPQA-M} & \textbf{GPQA-D} & \textbf{ACP} \\ 
\midrule
\multicolumn{9}{l}{\cellcolor{groupgray}\textit{Model: Qwen3-8B-Base}} \\
Base Model & 0.317 / 0.850 & 0.016 / 0.166 & 0.037 / 0.289 & 0.378 / 0.892 & 0.187 / 0.549 & 0.301 & 0.283 & 0.515 \\
GRPO & 0.713 / 0.949 & 0.160 / 0.344 & 0.228 / 0.497 & 0.782 / 0.890 & 0.471 / 0.670 & 0.353 & 0.384 & 0.562 \\
GRPO + Clip-High & 0.734 / 0.950 & 0.185 / 0.469 & 0.231 / 0.448 & 0.799 / 0.916 & 0.487 / 0.696 & 0.346 & 0.354 & 0.567 \\
\rowcolor{highlightgreen} 
\textbf{GRPO + ETR} & \textbf{0.790 / 0.967} & \textbf{0.195 / 0.482} & \textbf{0.254 / 0.532} & \textbf{0.819 / 0.926} & \textbf{0.515 / 0.727} & \textbf{0.393} & \textbf{0.444} & \textbf{0.590} \\
\midrule
\multicolumn{9}{l}{\cellcolor{groupgray}\textit{Model: Llama-3.1-8B-Instruct}} \\
Base Model & 0.070 / 0.486 & 0.029 / 0.168 & 0.003 / 0.063 & 0.059 / 0.403 & 0.040 / 0.280 & 0.281 & 0.227 & 0.464 \\
GRPO & 0.161 / 0.538 & \textbf{0.051} / 0.165 & 0.007 / 0.122 & 0.073 / 0.336 & 0.073 / 0.290 & 0.324 & 0.333 & 0.478 \\
GRPO + Clip-High & 0.170 / 0.640 & 0.038 / 0.186 & 0.008 / 0.157 & 0.088 / 0.449 & 0.076 / 0.358 & 0.326 & 0.313 & 0.485 \\
\rowcolor{highlightgreen}
\textbf{GRPO + ETR} & \textbf{0.178 / 0.693} & 0.049 / \textbf{0.202} & \textbf{0.016 / 0.203} & \textbf{0.203 / 0.604} & \textbf{0.112 / 0.426} & \textbf{0.339} & \textbf{0.354} & \textbf{0.516} \\
\midrule
\multicolumn{9}{l}{\cellcolor{groupgray}\textit{Model: Qwen2.5-7B-Math-Base}} \\
Base Model & 0.233 / 0.797 & 0.002 / 0.042 & 0.000 / 0.000 & 0.349 / 0.870 & 0.146 / 0.427 & 0.250 & 0.253 & 0.428 \\
GRPO & 0.442 / 0.829 & 0.204 / 0.360 & 0.119 / 0.295 & 0.586 / 0.885 & 0.338 / 0.592 & \textbf{0.275} & 0.278 & 0.437 \\
GRPO + Clip-High & 0.605 / 0.919 & 0.199 / 0.316 & \textbf{0.168} / 0.370 & 0.677 / \textbf{0.914} & 0.412 / 0.630 & 0.266 & 0.268 & 0.443 \\
\rowcolor{highlightgreen}
\textbf{GRPO + ETR} & \textbf{0.622 / 0.925} & \textbf{0.215 / 0.358} & 0.141 / \textbf{0.376} & \textbf{0.717} / 0.911 & \textbf{0.424 / 0.643} & 0.257 & \textbf{0.323} & \textbf{0.445} \\
\bottomrule
\end{tabular}
\end{table*}

\subsection{Analysis of Training Process}

We analyze the evolution of accuracy and policy entropy during training to understand the mechanism behind the performance gains.

\begin{figure}[t]
    \centering
    \includegraphics[width=\linewidth]{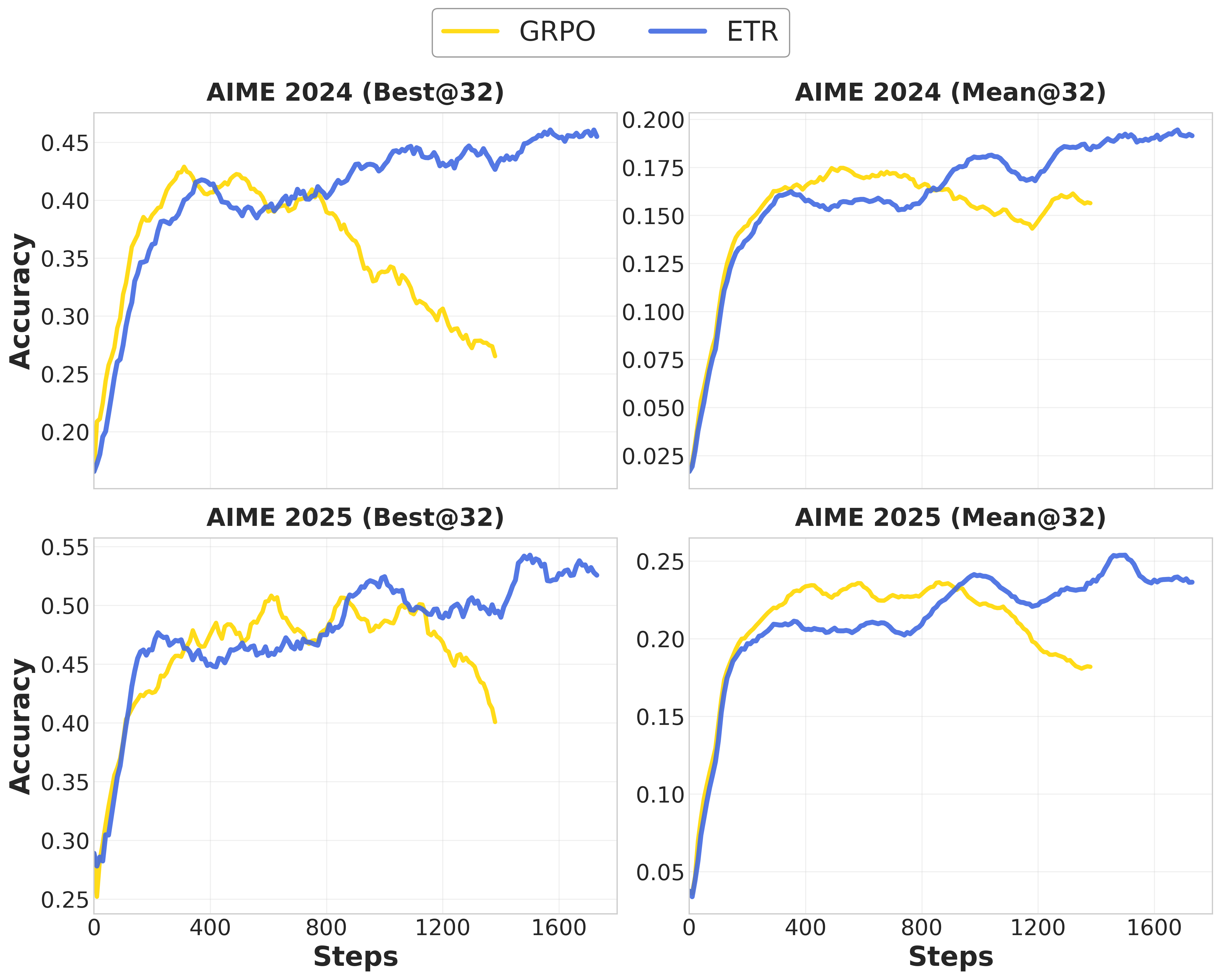}
    \caption{Val Accuracy on AIME 2024/2025 using Qwen3-8B-Base. Standard GRPO (yellow) suffers from performance collapse in later learning stages. ETR (blue) maintains a steady upwards trend in both Mean@32 and Best@32.}
    \label{fig:acc_curves}
\end{figure}

\paragraph{Mitigating Late-Stage Collapse.}
Figure~\ref{fig:acc_curves} illustrates the validation accuracy on AIME 2024. Standard GRPO exhibits significant instability in the later stages of training. After reaching a peak, the Best@32 metric begins to decline. We attribute this not to traditional overfitting, but to entropy collapse. In the multi-sampling setting ($G=32$), if the policy becomes deterministic, the diversity of generated responses vanishes. Once the model converges to a sub-optimal path, it loses the capability to explore and sample correct answers, leading to the degradation of the Best@32 metric. In contrast, ETR (Blue line) maintains a robust upward trajectory without collapse. This proves that the dynamic boundary mechanism effectively stabilizes the policy update throughout the training process.

\begin{figure}[t]
    \centering
    \includegraphics[width=\linewidth]{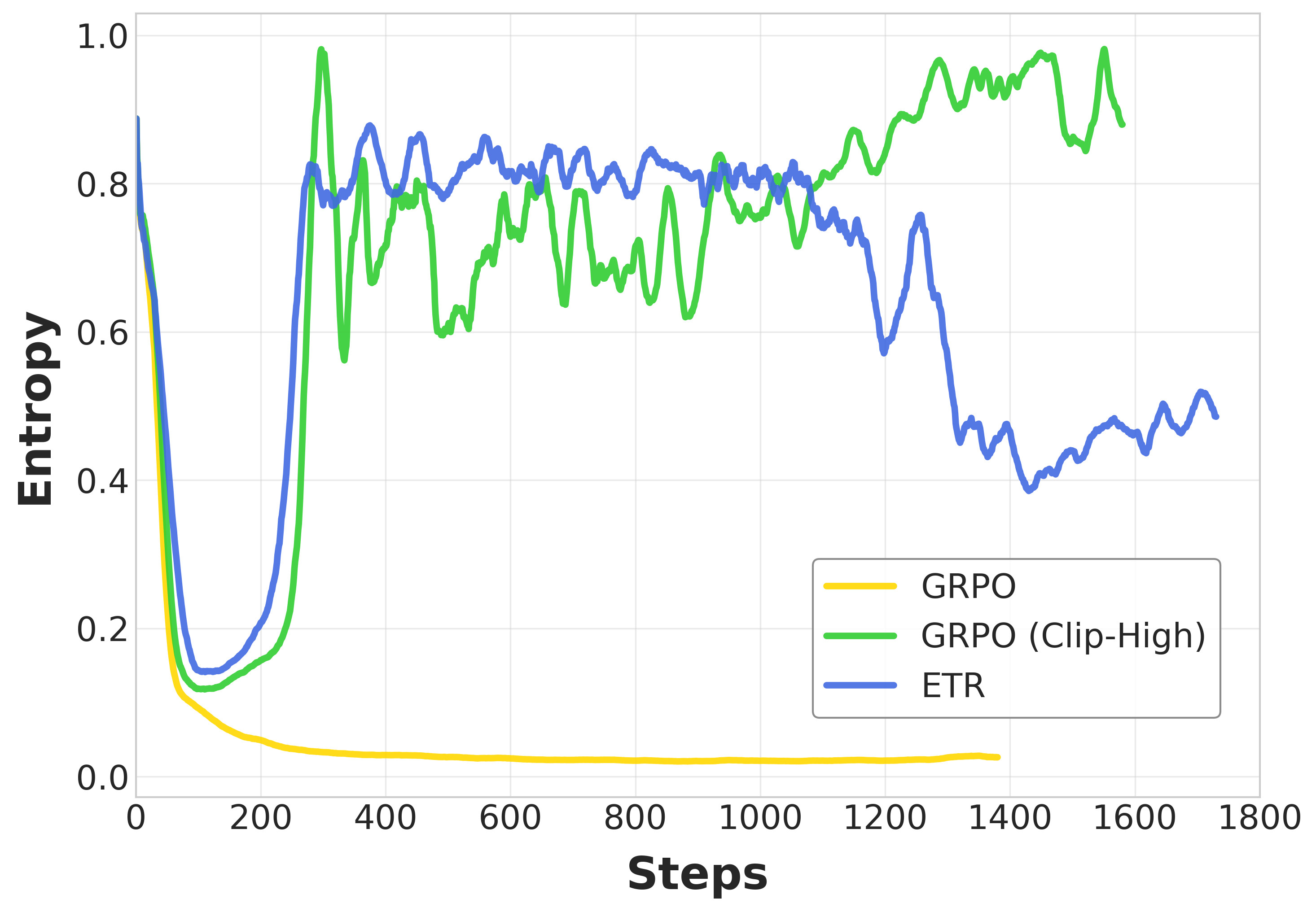}
    \caption{\textbf{Policy Entropy Evolution.} GRPO collapses to zero, limiting exploration. Clip-High leads to high, unstable entropy. ETR maintains healthy entropy levels, facilitating sustained learning.}
    \label{fig:entropy}
\end{figure}

\paragraph{Entropy Evolution.}
Figure~\ref{fig:entropy} depicts the evolution of policy entropy.

\noindent \textbf{GRPO:} The entropy drops rapidly to near-zero, indicating premature convergence to a deterministic distribution.
 
\noindent \textbf{GRPO-Clip-High:} Increasing the static threshold to 0.28 prevents entropy collapse but leads to severe oscillations and a continuous rise in entropy in later stages. This suggests that unconstrained exploration introduces harmful noise that does not translate into performance gains.
 
\noindent \textbf{ETR:} ETR exhibits a "dip-and-rebound" pattern. Entropy decreases initially as the model learns, but stabilizes or recovers in later stages. This resilience indicates that the elastic boundary allows the model to re-expand its search space when high-value signals are encountered, maintaining a balance between exploitation and exploration.

\subsection{Ablation Study}

We verify the design choices of ETR through two sets of ablation studies.

\begin{figure*}[t]
    \centering
    \includegraphics[width=0.9\linewidth]{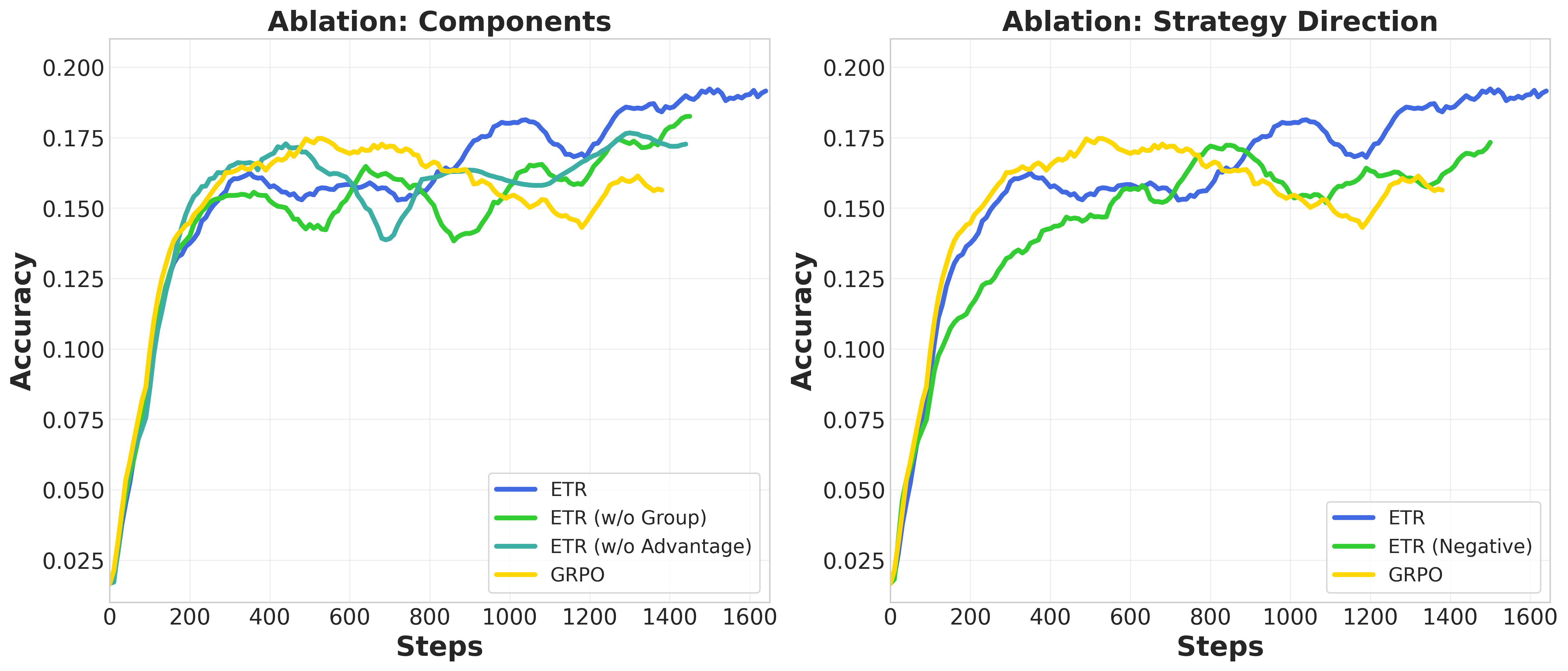}
    
    \vspace{-0.3cm} 
    
    \caption{\textbf{Ablation Study.} \textbf{Left:} Contribution of Macro and Micro components. \textbf{Right:} Comparison of inverse directions. The Negative Variant (expanding boundaries for negative samples) fails due to noise, confirming the necessity of our asymmetric design.}
    \label{fig:ablation}

\end{figure*}

\paragraph{Component Effectiveness.}
We compare the full ETR with variants where either the micro-level (sample-based) or macro-level (group-based) adjustment is disabled. Removing the micro-level adjustment ($\lambda_1=0$) results in the most significant performance drop and slower convergence. This confirms that adapting to the advantage magnitude is the primary driver of ETR, as static boundaries limit the utilization of high-value gradients. Removing the macro-level adjustment ($\lambda_2=0$) also degrades efficiency. This validates the effectiveness of implicit curriculum learning: allocating a larger update budget to groups with high variance (i.e., moderate difficulty) accelerates learning from the most informative batches.

\paragraph{Strategy Direction.}
We validate the necessity of the asymmetric design (expanding for positive samples, tightening for negative ones) by comparing ETR against an Inverse Variant (expanding for negative, tightening for positive). The results show that the Inverse Variant significantly underperforms. This is due to the asymmetry of gradient updates: increasing the probability of a correct token is a focused operation that benefits from a relaxed bound. In contrast, decreasing the probability of an error token redistributes probability mass to the rest of the vocabulary. If the constraint is too loose for negative samples, this redistribution indiscriminately boosts irrelevant tokens, introducing diffusion noise. Therefore, tightening the bound for negative samples is essential for stability.

\subsection{Computational Efficiency}

A significant advantage of ETR is its zero computational overhead relative to GRPO. Unlike methods that require auxiliary models or complex matrix computations, ETR relies solely on element-wise operations on tensors that are already computed during the standard rollout and advantage estimation phases. The calculation of the dynamic threshold adds negligible latency to the training step, making ETR highly scalable and easy to integrate into existing RLVR pipelines.

\subsection{Hyperparameter Robustness}

Finally, we examine the sensitivity of ETR to hyperparameters. Experimental results indicate that setting both the micro and macro coefficients to 0.1 is a robust choice. We applied this identical configuration across different model architectures (Qwen and Llama) and sizes, and consistently achieved performance gains over the baseline. This suggests that the effectiveness of ETR stems from its core signal-aware mechanism rather than overfitting to a specific set of hyperparameters.
\section{Related Works}

\subsection{Reinforcement Learning for LLM Post-Training}
Reinforcement learning has become a central paradigm for post-training large language models, most prominently through Reinforcement Learning from Human Feedback (RLHF)~\cite{ouyang2022training,rlhfsurvey,onlinerlhf}, where policies are optimized using learned reward models. While effective for alignment, reward-model-based approaches introduce additional approximation errors and often struggle to provide precise supervision for tasks with objectively verifiable outcomes. This has motivated the recent shift toward Reinforcement Learning with Verifiable Rewards (RLVR)~\cite{wen2025reinforcement,tian2025seea}, particularly in domains such as mathematical reasoning and program synthesis, where correctness can be automatically evaluated~\cite{TrainingVerifiers,letsverifystepstep}. We focus on improving the stability and efficiency of policy optimization under outcome-driven rewards in RLVR.

\subsection{Trust-Region and PPO-Style Policy Optimization}

Trust-region methods are widely used to stabilize policy optimization. TRPO~\cite{schulman2017trustregionpolicyoptimization} enforces an explicit constraint on policy updates via KL divergence, while PPO~\cite{schulman2017proximal} introduces a clipped surrogate objective as a computationally efficient approximation. PPO-style clipping has become the de facto standard in RLHF and RLVR pipelines due to its simplicity and robustness~\cite{bai2022constitutionalaiharmlessnessai,gpto1}. However, this design relies on a static trust region that applies uniformly across samples, implicitly assuming homogeneous training signals. While prior work has explored alternative constraints and adaptive penalties~\cite{adaptive1,su2025trustregionadaptivepolicyoptimization,yu2025dapo}, the suitability of static trust regions for heterogeneous, outcome-driven learning remains underexplored. Our work revisits this assumption and proposes a data-adaptive alternative.

\subsection{Adaptive Optimization and Curriculum Learning}

Adapting optimization behavior to data difficulty and information content has been widely studied in reinforcement learning and optimization. Prior works~\cite{schulman2017trustregionpolicyoptimization,AdaptiveOptimization,AdaGC} explore adaptive learning rates, trust-region adjustments, and variance-aware update rules to improve stability and sample efficiency. In parallel, curriculum learning~\cite{bengio2009curriculum,curriculumsurvey,Okamoto2021ReinforcementLW} methods aim to prioritize informative or appropriately challenging data, either through manually designed schedules or learned difficulty estimators~\cite{song2025fastcurl,shi2025efficient,zhang2025clpocurriculumlearningmeets}. While effective, many curriculum strategies rely on task-specific heuristics or additional supervision that may not be readily available in large-scale RL settings.
In the context of LLM post-training, GRPO can be viewed as an implicit step toward difficulty-aware learning by computing relative advantages within groups, reducing reliance on absolute reward scales. However, GRPO still applies a uniform trust region across samples and groups, limiting its ability to adapt optimization dynamics to heterogeneous outcome statistics. Our work builds on this line of research by introducing Elastic Trust Regions, which directly leverage advantage magnitude and group-level variance to dynamically adjust optimization constraints, enabling implicit curriculum learning without additional supervision.
\vspace{0.12cm}
\section{Conclusion}
\vspace{0.13cm}
We identify static trust regions as a key limitation of GRPO when applied to heterogeneous, outcome-driven reasoning tasks. To address this, we propose Elastic Trust Regions (ETR), a data-adaptive constraint mechanism that adjusts policy updates based on advantage magnitude and group-level variance. Experiments on mathematical reasoning benchmarks show that ETR improves accuracy and stabilizes training by mitigating policy entropy collapse. These results highlight the importance of outcome-adaptive optimization for effective post-training of reasoning-oriented language models.
Given its simplicity and negligible computational overhead, ETR can serve as a plug-and-play enhancement for existing RLVR pipelines. Ultimately, bridging signal quality and constraints enables robust, sample-efficient training for reasoning models and their future large-scale applications.

\bibliography{custom}

\begin{thebibliography}{42}
\providecommand{\natexlab}[1]{#1}

\bibitem[{Achiam et~al.(2023)Achiam, Adler, Agarwal, Ahmad, Akkaya, Aleman, Almeida, Altenschmidt, Altman, Anadkat et~al.}]{achiam2023gpt}
Josh Achiam, Steven Adler, Sandhini Agarwal, Lama Ahmad, Ilge Akkaya, Florencia~Leoni Aleman, Diogo Almeida, Janko Altenschmidt, Sam Altman, Shyamal Anadkat, and 1 others. 2023.
\newblock Gpt-4 technical report.
\newblock \emph{arXiv preprint arXiv:2303.08774}.

\bibitem[{Bengio et~al.(2009)Bengio, Louradour, Collobert, and Weston}]{bengio2009curriculum}
Yoshua Bengio, J{\'e}r{\^o}me Louradour, Ronan Collobert, and Jason Weston. 2009.
\newblock Curriculum learning.
\newblock In \emph{Proceedings of the 26th annual international conference on machine learning}, pages 41--48.

\bibitem[{Cobbe et~al.(2021)Cobbe, Kosaraju, Bavarian, Chen, Jun, Kaiser, Plappert, Tworek, Hilton, Nakano, Hesse, and Schulman}]{TrainingVerifiers}
Karl Cobbe, Vineet Kosaraju, Mohammad Bavarian, Mark Chen, Heewoo Jun, Lukasz Kaiser, Matthias Plappert, Jerry Tworek, Jacob Hilton, Reiichiro Nakano, Christopher Hesse, and John Schulman. 2021.
\newblock \href {https://arxiv.org/abs/2110.14168} {Training verifiers to solve math word problems}.
\newblock \emph{CoRR}, abs/2110.14168.

\bibitem[{Dong et~al.(2024)Dong, Xiong, Pang, Wang, Zhao, Zhou, Jiang, Sahoo, Xiong, and Zhang}]{onlinerlhf}
Hanze Dong, Wei Xiong, Bo~Pang, Haoxiang Wang, Han Zhao, Yingbo Zhou, Nan Jiang, Doyen Sahoo, Caiming Xiong, and Tong Zhang. 2024.
\newblock \href {https://arxiv.org/abs/2405.07863} {Rlhf workflow: From reward modeling to online rlhf}.
\newblock \emph{Preprint}, arXiv:2405.07863.

\bibitem[{Grattafiori et~al.(2024)Grattafiori, Dubey, Jauhri, Pandey, Kadian, Al-Dahle, Letman, Mathur, Schelten, Vaughan, Yang, Fan, Goyal, Hartshorn, Yang, Mitra, Sravankumar, Korenev, Hinsvark, Rao, Zhang, Rodriguez, Gregerson, Spataru, Roziere, Biron, Tang, Chern, Caucheteux, Nayak, Bi, Marra, McConnell, Keller, Touret, Wu, Wong, Ferrer, Nikolaidis, Allonsius, Song, Pintz, Livshits, Wyatt, Esiobu, Choudhary, Mahajan, Garcia-Olano, Perino, Hupkes, Lakomkin, AlBadawy, Lobanova, Dinan, Smith, Radenovic, Guzmán, Zhang, Synnaeve, Lee, Anderson, Thattai, Nail, Mialon, Pang, Cucurell, Nguyen, Korevaar, Xu, Touvron, Zarov, Ibarra, Kloumann, Misra, Evtimov, Zhang, Copet, Lee, Geffert, Vranes, Park, Mahadeokar, Shah, van~der Linde, Billock, Hong, Lee, Fu, Chi, Huang, Liu, Wang, Yu, Bitton, Spisak, Park, Rocca, Johnstun, Saxe, Jia, Alwala, Prasad, Upasani, Plawiak, Li, Heafield, Stone, El-Arini, Iyer, Malik, Chiu, Bhalla, Lakhotia, Rantala-Yeary, van~der Maaten, Chen, Tan, Jenkins, Martin, Madaan, Malo, Blecher,
  Landzaat, de~Oliveira, Muzzi, Pasupuleti, Singh, Paluri, Kardas, Tsimpoukelli, Oldham, Rita, Pavlova, Kambadur, Lewis, Si, Singh, Hassan, Goyal, Torabi, Bashlykov, Bogoychev, Chatterji, Zhang, Duchenne, Çelebi, Alrassy, Zhang, Li, Vasic, Weng, Bhargava, Dubal, Krishnan, Koura, Xu, He, Dong, Srinivasan, Ganapathy, Calderer, Cabral, Stojnic, Raileanu, Maheswari, Girdhar, Patel, Sauvestre, Polidoro, Sumbaly, Taylor, Silva, Hou, Wang, Hosseini, Chennabasappa, Singh, Bell, Kim, Edunov, Nie, Narang, Raparthy, Shen, Wan, Bhosale, Zhang, Vandenhende, Batra, Whitman, Sootla, Collot, Gururangan, Borodinsky, Herman, Fowler, Sheasha, Georgiou, Scialom, Speckbacher, Mihaylov, Xiao, Karn, Goswami, Gupta, Ramanathan, Kerkez, Gonguet, Do, Vogeti, Albiero, Petrovic, Chu, Xiong, Fu, Meers, Martinet, Wang, Wang, Tan, Xia, Xie, Jia, Wang, Goldschlag, Gaur, Babaei, Wen, Song, Zhang, Li, Mao, Coudert, Yan, Chen, Papakipos, Singh, Srivastava, Jain, Kelsey, Shajnfeld, Gangidi, Victoria, Goldstand, Menon, Sharma, Boesenberg,
  Baevski, Feinstein, Kallet, Sangani, Teo, Yunus, Lupu, Alvarado, Caples, Gu, Ho, Poulton, Ryan, Ramchandani, Dong, Franco, Goyal, Saraf, Chowdhury, Gabriel, Bharambe, Eisenman, Yazdan, James, Maurer, Leonhardi, Huang, Loyd, Paola, Paranjape, Liu, Wu, Ni, Hancock, Wasti, Spence, Stojkovic, Gamido, Montalvo, Parker, Burton, Mejia, Liu, Wang, Kim, Zhou, Hu, Chu, Cai, Tindal, Feichtenhofer, Gao, Civin, Beaty, Kreymer, Li, Adkins, Xu, Testuggine, David, Parikh, Liskovich, Foss, Wang, Le, Holland, Dowling, Jamil, Montgomery, Presani, Hahn, Wood, Le, Brinkman, Arcaute, Dunbar, Smothers, Sun, Kreuk, Tian, Kokkinos, Ozgenel, Caggioni, Kanayet, Seide, Florez, Schwarz, Badeer, Swee, Halpern, Herman, Sizov, Guangyi, Zhang, Lakshminarayanan, Inan, Shojanazeri, Zou, Wang, Zha, Habeeb, Rudolph, Suk, Aspegren, Goldman, Zhan, Damlaj, Molybog, Tufanov, Leontiadis, Veliche, Gat, Weissman, Geboski, Kohli, Lam, Asher, Gaya, Marcus, Tang, Chan, Zhen, Reizenstein, Teboul, Zhong, Jin, Yang, Cummings, Carvill, Shepard, McPhie,
  Torres, Ginsburg, Wang, Wu, U, Saxena, Khandelwal, Zand, Matosich, Veeraraghavan, Michelena, Li, Jagadeesh, Huang, Chawla, Huang, Chen, Garg, A, Silva, Bell, Zhang, Guo, Yu, Moshkovich, Wehrstedt, Khabsa, Avalani, Bhatt, Mankus, Hasson, Lennie, Reso, Groshev, Naumov, Lathi, Keneally, Liu, Seltzer, Valko, Restrepo, Patel, Vyatskov, Samvelyan, Clark, Macey, Wang, Hermoso, Metanat, Rastegari, Bansal, Santhanam, Parks, White, Bawa, Singhal, Egebo, Usunier, Mehta, Laptev, Dong, Cheng, Chernoguz, Hart, Salpekar, Kalinli, Kent, Parekh, Saab, Balaji, Rittner, Bontrager, Roux, Dollar, Zvyagina, Ratanchandani, Yuvraj, Liang, Alao, Rodriguez, Ayub, Murthy, Nayani, Mitra, Parthasarathy, Li, Hogan, Battey, Wang, Howes, Rinott, Mehta, Siby, Bondu, Datta, Chugh, Hunt, Dhillon, Sidorov, Pan, Mahajan, Verma, Yamamoto, Ramaswamy, Lindsay, Lindsay, Feng, Lin, Zha, Patil, Shankar, Zhang, Zhang, Wang, Agarwal, Sajuyigbe, Chintala, Max, Chen, Kehoe, Satterfield, Govindaprasad, Gupta, Deng, Cho, Virk, Subramanian, Choudhury,
  Goldman, Remez, Glaser, Best, Koehler, Robinson, Li, Zhang, Matthews, Chou, Shaked, Vontimitta, Ajayi, Montanez, Mohan, Kumar, Mangla, Ionescu, Poenaru, Mihailescu, Ivanov, Li, Wang, Jiang, Bouaziz, Constable, Tang, Wu, Wang, Wu, Gao, Kleinman, Chen, Hu, Jia, Qi, Li, Zhang, Zhang, Adi, Nam, Yu, Wang, Zhao, Hao, Qian, Li, He, Rait, DeVito, Rosnbrick, Wen, Yang, Zhao, and Ma}]{grattafiori2024llama3herdmodels}
Aaron Grattafiori, Abhimanyu Dubey, Abhinav Jauhri, Abhinav Pandey, Abhishek Kadian, Ahmad Al-Dahle, Aiesha Letman, Akhil Mathur, Alan Schelten, Alex Vaughan, Amy Yang, Angela Fan, Anirudh Goyal, Anthony Hartshorn, Aobo Yang, Archi Mitra, Archie Sravankumar, Artem Korenev, Arthur Hinsvark, and 542 others. 2024.
\newblock \href {https://arxiv.org/abs/2407.21783} {The llama 3 herd of models}.
\newblock \emph{Preprint}, arXiv:2407.21783.

\bibitem[{Guo et~al.(2025)Guo, Yang, Zhang, Song, Zhang, Xu, Zhu, Ma, Wang, Bi et~al.}]{guo2025deepseek}
Daya Guo, Dejian Yang, Haowei Zhang, Junxiao Song, Ruoyu Zhang, Runxin Xu, Qihao Zhu, Shirong Ma, Peiyi Wang, Xiao Bi, and 1 others. 2025.
\newblock Deepseek-r1: Incentivizing reasoning capability in llms via reinforcement learning.
\newblock \emph{arXiv preprint arXiv:2501.12948}.

\bibitem[{Hanzely(2023)}]{AdaptiveOptimization}
Slavomír Hanzely. 2023.
\newblock \href {https://arxiv.org/abs/2311.10203} {Adaptive optimization algorithms for machine learning}.
\newblock \emph{Preprint}, arXiv:2311.10203.

\bibitem[{Hendrycks et~al.(2021)Hendrycks, Burns, Kadavath, Arora, Basart, Tang, Song, and Steinhardt}]{hendrycks2021measuring}
Dan Hendrycks, Collin Burns, Saurav Kadavath, Akul Arora, Steven Basart, Eric Tang, Dawn Song, and Jacob Steinhardt. 2021.
\newblock Measuring mathematical problem solving with the math dataset.
\newblock \emph{arXiv preprint arXiv:2103.03874}.

\bibitem[{Jaech et~al.(2024)Jaech, Kalai, Lerer, Richardson, El-Kishky, Low, Helyar, Madry, Beutel, Carney et~al.}]{jaech2024openai}
Aaron Jaech, Adam Kalai, Adam Lerer, Adam Richardson, Ahmed El-Kishky, Aiden Low, Alec Helyar, Aleksander Madry, Alex Beutel, Alex Carney, and 1 others. 2024.
\newblock Openai o1 system card.
\newblock \emph{arXiv preprint arXiv:2412.16720}.

\bibitem[{Kokel et~al.(2025)Kokel, Katz, Srinivas, and Sohrabi}]{kokel2024acp}
Harsha Kokel, Michael Katz, Kavitha Srinivas, and Shirin Sohrabi. 2025.
\newblock Acpbench: Reasoning about action, change, and planning.
\newblock In \emph{{AAAI}}. {AAAI} Press.

\bibitem[{Lambert(2026)}]{rlhfsurvey}
Nathan Lambert. 2026.
\newblock \href {https://arxiv.org/abs/2504.12501} {Reinforcement learning from human feedback}.
\newblock \emph{Preprint}, arXiv:2504.12501.

\bibitem[{Lewkowycz et~al.(2022)Lewkowycz, Andreassen, Dohan, Dyer, Michalewski, Ramasesh, Slone, Anil, Schlag, Gutman-Solo et~al.}]{lewkowycz2022solving}
Aitor Lewkowycz, Anders Andreassen, David Dohan, Ethan Dyer, Henryk Michalewski, Vinay Ramasesh, Ambrose Slone, Cem Anil, Imanol Schlag, Theo Gutman-Solo, and 1 others. 2022.
\newblock Solving quantitative reasoning problems with language models.
\newblock \emph{Advances in neural information processing systems}, 35:3843--3857.

\bibitem[{Li et~al.(2024)Li, Beeching, Tunstall, Lipkin, Soletskyi, Huang, Rasul, Yu, Jiang, Shen et~al.}]{li2024numinamath}
Jia Li, Edward Beeching, Lewis Tunstall, Ben Lipkin, Roman Soletskyi, Shengyi Huang, Kashif Rasul, Longhui Yu, Albert~Q Jiang, Ziju Shen, and 1 others. 2024.
\newblock Numinamath: The largest public dataset in ai4maths with 860k pairs of competition math problems and solutions.
\newblock \emph{Hugging Face repository}, 13(9):9.

\bibitem[{Lightman et~al.(2023)Lightman, Kosaraju, Burda, Edwards, Baker, Lee, Leike, Schulman, Sutskever, and Cobbe}]{letsverifystepstep}
Hunter Lightman, Vineet Kosaraju, Yura Burda, Harri Edwards, Bowen Baker, Teddy Lee, Jan Leike, John Schulman, Ilya Sutskever, and Karl Cobbe. 2023.
\newblock \href {https://arxiv.org/abs/2305.20050} {Let's verify step by step}.
\newblock \emph{Preprint}, arXiv:2305.20050.

\bibitem[{Lin et~al.(2025)Lin, Mi, and Gao}]{curriculumsurvey}
Siyu Lin, Qingwei Mi, and Tianhan Gao. 2025.
\newblock \href {https://doi.org/10.1109/CCWC62904.2025.10903795} {A survey of curriculum learning in deep reinforcement learning}.
\newblock In \emph{2025 IEEE 15th Annual Computing and Communication Workshop and Conference (CCWC)}, pages 01141--01147.

\bibitem[{Okamoto et~al.(2021)Okamoto, Kera, and Kawamoto}]{Okamoto2021ReinforcementLW}
Wataru Okamoto, Hiroshi Kera, and K.~Kawamoto. 2021.
\newblock \href {https://api.semanticscholar.org/CorpusID:244463103} {Reinforcement learning with adaptive curriculum dynamics randomization for fault-tolerant robot control}.
\newblock \emph{ArXiv}, abs/2111.10005.

\bibitem[{Ouyang et~al.(2022)Ouyang, Wu, Jiang, Almeida, Wainwright, Mishkin, Zhang, Agarwal, Slama, Ray et~al.}]{ouyang2022training}
Long Ouyang, Jeffrey Wu, Xu~Jiang, Diogo Almeida, Carroll Wainwright, Pamela Mishkin, Chong Zhang, Sandhini Agarwal, Katarina Slama, Alex Ray, and 1 others. 2022.
\newblock Training language models to follow instructions with human feedback.
\newblock \emph{Advances in neural information processing systems}, 35:27730--27744.

\bibitem[{Rein et~al.(2024)Rein, Hou, Stickland, Petty, Pang, Dirani, Michael, and Bowman}]{rein2024gpqa}
David Rein, Betty~Li Hou, Asa~Cooper Stickland, Jackson Petty, Richard~Yuanzhe Pang, Julien Dirani, Julian Michael, and Samuel~R Bowman. 2024.
\newblock Gpqa: A graduate-level google-proof q\&a benchmark.
\newblock In \emph{First Conference on Language Modeling}.

\bibitem[{Schulman et~al.(2017{\natexlab{a}})Schulman, Levine, Moritz, Jordan, and Abbeel}]{schulman2017trustregionpolicyoptimization}
John Schulman, Sergey Levine, Philipp Moritz, Michael~I. Jordan, and Pieter Abbeel. 2017{\natexlab{a}}.
\newblock \href {https://arxiv.org/abs/1502.05477} {Trust region policy optimization}.
\newblock \emph{Preprint}, arXiv:1502.05477.

\bibitem[{Schulman et~al.(2015)Schulman, Moritz, Levine, Jordan, and Abbeel}]{schulman2015high}
John Schulman, Philipp Moritz, Sergey Levine, Michael Jordan, and Pieter Abbeel. 2015.
\newblock High-dimensional continuous control using generalized advantage estimation.
\newblock \emph{arXiv preprint arXiv:1506.02438}.

\bibitem[{Schulman et~al.(2017{\natexlab{b}})Schulman, Wolski, Dhariwal, Radford, and Klimov}]{schulman2017proximal}
John Schulman, Filip Wolski, Prafulla Dhariwal, Alec Radford, and Oleg Klimov. 2017{\natexlab{b}}.
\newblock Proximal policy optimization algorithms.
\newblock \emph{arXiv preprint arXiv:1707.06347}.

\bibitem[{Shao et~al.(2024)Shao, Wang, Zhu, Xu, Song, Bi, Zhang, Zhang, Li, Wu et~al.}]{shao2024deepseekmath}
Zhihong Shao, Peiyi Wang, Qihao Zhu, Runxin Xu, Junxiao Song, Xiao Bi, Haowei Zhang, Mingchuan Zhang, YK~Li, Yang Wu, and 1 others. 2024.
\newblock Deepseekmath: Pushing the limits of mathematical reasoning in open language models.
\newblock \emph{arXiv preprint arXiv:2402.03300}.

\bibitem[{Sheng et~al.(2025)Sheng, Zhang, Ye, Wu, Zhang, Zhang, Peng, Lin, and Wu}]{sheng2025hybridflow}
Guangming Sheng, Chi Zhang, Zilingfeng Ye, Xibin Wu, Wang Zhang, Ru~Zhang, Yanghua Peng, Haibin Lin, and Chuan Wu. 2025.
\newblock Hybridflow: A flexible and efficient rlhf framework.
\newblock In \emph{Proceedings of the Twentieth European Conference on Computer Systems}, pages 1279--1297.

\bibitem[{Shi et~al.(2025)Shi, Wu, Song, Zhou, and Zhao}]{shi2025efficient}
Taiwei Shi, Yiyang Wu, Linxin Song, Tianyi Zhou, and Jieyu Zhao. 2025.
\newblock Efficient reinforcement finetuning via adaptive curriculum learning.
\newblock \emph{arXiv preprint arXiv:2504.05520}.

\bibitem[{Song et~al.(2025)Song, Zheng, Li, Yang, Luo, Pan, and Zhang}]{song2025fastcurl}
Mingyang Song, Mao Zheng, Zheng Li, Wenjie Yang, Xuan Luo, Yue Pan, and Feng Zhang. 2025.
\newblock Fastcurl: Curriculum reinforcement learning with stage-wise context scaling for efficient training r1-like reasoning models.
\newblock \emph{arXiv preprint arXiv:2503.17287}.

\bibitem[{Su et~al.(2025)Su, Guan, Gu, Huang, and Wang}]{su2025trustregionadaptivepolicyoptimization}
Mingyu Su, Jian Guan, Yuxian Gu, Minlie Huang, and Hongning Wang. 2025.
\newblock \href {https://arxiv.org/abs/2512.17636} {Trust-region adaptive policy optimization}.
\newblock \emph{Preprint}, arXiv:2512.17636.

\bibitem[{Team(2022)}]{bai2022constitutionalaiharmlessnessai}
Constitutional~AI Team. 2022.
\newblock \href {https://arxiv.org/abs/2212.08073} {Constitutional ai: Harmlessness from ai feedback}.
\newblock \emph{Preprint}, arXiv:2212.08073.

\bibitem[{Team(2025)}]{gemini2.5}
Gemini Team. 2025.
\newblock \href {https://arxiv.org/abs/2507.06261} {Gemini 2.5: Pushing the frontier with advanced reasoning, multimodality, long context, and next generation agentic capabilities}.
\newblock \emph{Preprint}, arXiv:2507.06261.

\bibitem[{Team et~al.(2025)Team, Du, Gao, Xing, Jiang, Chen, Li, Xiao, Du, Liao, Tang, Wang, Zhang, Yuan, Lu, Tang, Sung, Wei, Lai, Guo, Zhu, Ding, Hu, Yang, Zhang, Yao, Zhao, Lu, Li, Yu, Gao, Zheng, Yuan, Chen, Guo, Su, Wang, Zhao, Zhang, Liu, Yan, Wu, Shi, Ye, Yu, Dong, Zhang, Ma, Pan, Gong, Liu, Ma, Wei, Cao, Huang, Jiang, Gao, Xiong, He, Huang, Xu, Wu, He, Wei, Jia, Wu, Xu, Zu, Zhou, Pan, Charles, Li, Hu, Liu, Chen, Wang, Liu, Qin, Liu, Yang, Bao, Du, Wu, Wang, Zhou, Wang, Li, Zhu, Zhang, Wang, Yang, Huang, Huang, Xu, Yang, and Lin}]{kimiteam2025kimik15scalingreinforcement}
Kimi Team, Angang Du, Bofei Gao, Bowei Xing, Changjiu Jiang, Cheng Chen, Cheng Li, Chenjun Xiao, Chenzhuang Du, Chonghua Liao, Chuning Tang, Congcong Wang, Dehao Zhang, Enming Yuan, Enzhe Lu, Fengxiang Tang, Flood Sung, Guangda Wei, Guokun Lai, and 77 others. 2025.
\newblock \href {https://arxiv.org/abs/2501.12599} {Kimi k1.5: Scaling reinforcement learning with llms}.
\newblock \emph{Preprint}, arXiv:2501.12599.

\bibitem[{Team(2024)}]{gpto1}
OpenAI Team. 2024.
\newblock \href {https://arxiv.org/abs/2412.16720} {Openai o1 system card}.
\newblock \emph{Preprint}, arXiv:2412.16720.

\bibitem[{Tian et~al.(2025)Tian, Zhang, Zhang, Chi, Luo, Lu, Fan, Zhou, Zhao, Lin et~al.}]{tian2025seea}
Wanxin Tian, Shijie Zhang, Kevin Zhang, Xiaowei Chi, Yulin Luo, Junyu Lu, Chunkai Fan, Qiang Zhou, Yiming Zhao, Ning Liu~Siyu Lin, and 1 others. 2025.
\newblock Seea-r1: Tree-structured reinforcement fine-tuning for self-evolving embodied agents.
\newblock \emph{arXiv preprint arXiv:2506.21669}.

\bibitem[{Touvron et~al.(2023)Touvron, Martin, Stone, Albert, Almahairi, Babaei, Bashlykov, Batra, Bhargava, Bhosale et~al.}]{touvron2023llama}
Hugo Touvron, Louis Martin, Kevin Stone, Peter Albert, Amjad Almahairi, Yasmine Babaei, Nikolay Bashlykov, Soumya Batra, Prajjwal Bhargava, Shruti Bhosale, and 1 others. 2023.
\newblock Llama 2: Open foundation and fine-tuned chat models.
\newblock \emph{arXiv preprint arXiv:2307.09288}.

\bibitem[{Wang et~al.(2024)Wang, Su, Dai, Lu, and Liu}]{AdaGC}
Qi~Wang, Feng Su, Shipeng Dai, Xiaojun Lu, and Yang Liu. 2024.
\newblock \href {https://doi.org/10.1016/j.eswa.2024.124956} {Adagc: A novel adaptive optimization algorithm with gradient bias correction}.
\newblock \emph{Expert Systems with Applications}, 256:124956.

\bibitem[{Wang et~al.(2025)Wang, Yang, Zeng, Ren, Liu, Peng, Cheng, He, Wang, Gao et~al.}]{wang2025reinforcement}
Yiping Wang, Qing Yang, Zhiyuan Zeng, Liliang Ren, Liyuan Liu, Baolin Peng, Hao Cheng, Xuehai He, Kuan Wang, Jianfeng Gao, and 1 others. 2025.
\newblock Reinforcement learning for reasoning in large language models with one training example.
\newblock \emph{arXiv preprint arXiv:2504.20571}.

\bibitem[{Wang et~al.(2019)Wang, He, Tan, and Gan}]{adaptive1}
Yuhui Wang, Hao He, Xiaoyang Tan, and Yaozhong Gan. 2019.
\newblock \emph{Trust region-guided proximal policy optimization}.
\newblock Curran Associates Inc., Red Hook, NY, USA.

\bibitem[{Wen et~al.(2025{\natexlab{a}})Wen, Liu, Zheng, Xu, Ye, Wu, Liang, Wang, Li, Miao et~al.}]{wen2025reinforcement}
Xumeng Wen, Zihan Liu, Shun Zheng, Zhijian Xu, Shengyu Ye, Zhirong Wu, Xiao Liang, Yang Wang, Junjie Li, Ziming Miao, and 1 others. 2025{\natexlab{a}}.
\newblock Reinforcement learning with verifiable rewards implicitly incentivizes correct reasoning in base llms.
\newblock \emph{arXiv preprint arXiv:2506.14245}.

\bibitem[{Wen et~al.(2025{\natexlab{b}})Wen, Liu, Zheng, Ye, Wu, Wang, Xu, Liang, Li, Miao, Bian, and Yang}]{wen2025reinforcementlearningverifiablerewards}
Xumeng Wen, Zihan Liu, Shun Zheng, Shengyu Ye, Zhirong Wu, Yang Wang, Zhijian Xu, Xiao Liang, Junjie Li, Ziming Miao, Jiang Bian, and Mao Yang. 2025{\natexlab{b}}.
\newblock \href {https://arxiv.org/abs/2506.14245} {Reinforcement learning with verifiable rewards implicitly incentivizes correct reasoning in base llms}.
\newblock \emph{Preprint}, arXiv:2506.14245.

\bibitem[{Yang et~al.(2025)Yang, Li, Yang, Zhang, Hui, Zheng, Yu, Gao, Huang, Lv et~al.}]{yang2025qwen3}
An~Yang, Anfeng Li, Baosong Yang, Beichen Zhang, Binyuan Hui, Bo~Zheng, Bowen Yu, Chang Gao, Chengen Huang, Chenxu Lv, and 1 others. 2025.
\newblock Qwen3 technical report.
\newblock \emph{arXiv preprint arXiv:2505.09388}.

\bibitem[{Yang et~al.(2024)Yang, Zhang, Hui, Gao, Yu, Li, Liu, Tu, Zhou, Lin, Lu, Xue, Lin, Liu, Ren, and Zhang}]{yang2024qwen25mathtechnicalreportmathematical}
An~Yang, Beichen Zhang, Binyuan Hui, Bofei Gao, Bowen Yu, Chengpeng Li, Dayiheng Liu, Jianhong Tu, Jingren Zhou, Junyang Lin, Keming Lu, Mingfeng Xue, Runji Lin, Tianyu Liu, Xingzhang Ren, and Zhenru Zhang. 2024.
\newblock \href {https://arxiv.org/abs/2409.12122} {Qwen2.5-math technical report: Toward mathematical expert model via self-improvement}.
\newblock \emph{Preprint}, arXiv:2409.12122.

\bibitem[{Yu et~al.(2025)Yu, Zhang, Zhu, Yuan, Zuo, Yue, Dai, Fan, Liu, Liu et~al.}]{yu2025dapo}
Qiying Yu, Zheng Zhang, Ruofei Zhu, Yufeng Yuan, Xiaochen Zuo, Yu~Yue, Weinan Dai, Tiantian Fan, Gaohong Liu, Lingjun Liu, and 1 others. 2025.
\newblock Dapo: An open-source llm reinforcement learning system at scale.
\newblock \emph{arXiv preprint arXiv:2503.14476}.

\bibitem[{Zhang et~al.(2025{\natexlab{a}})Zhang, Wang, Liu, Liu, Grabowski, Ie, Wang, and Li}]{zhang2025markovianreflectiveexplorationbayesadaptive}
Shenao Zhang, Yaqing Wang, Yinxiao Liu, Tianqi Liu, Peter Grabowski, Eugene Ie, Zhaoran Wang, and Yunxuan Li. 2025{\natexlab{a}}.
\newblock \href {https://arxiv.org/abs/2505.20561} {Beyond markovian: Reflective exploration via bayes-adaptive rl for llm reasoning}.
\newblock \emph{Preprint}, arXiv:2505.20561.

\bibitem[{Zhang et~al.(2025{\natexlab{b}})Zhang, Sun, Zhang, Guo, and Guo}]{zhang2025clpocurriculumlearningmeets}
Shijie Zhang, Guohao Sun, Kevin Zhang, Xiang Guo, and Rujun Guo. 2025{\natexlab{b}}.
\newblock \href {https://arxiv.org/abs/2509.25004} {Clpo: Curriculum learning meets policy optimization for llm reasoning}.
\newblock \emph{Preprint}, arXiv:2509.25004.

\end{thebibliography}

\clearpage

\appendix
\section{Theoretical Derivation}
\label{app:theoretical_derivation}

In this section, we provide the complete mathematical derivation of the Elastic Trust Regions (ETR) framework. We detail the transition from the theoretical weighted constraint to the practical dynamic clipping algorithm.

\subsection{Problem Formulation}

We consider the standard reinforcement learning objective with a trust region constraint. Let $\pi_{\theta}$ denote the current policy and $\pi_{\theta_{old}}$ denote the behavior policy. The probability ratio is defined as $r_t(\theta) = \frac{\pi_{\theta}(a_t|s_t)}{\pi_{\theta_{old}}(a_t|s_t)}$.

Standard Trust Region Policy Optimization (TRPO) imposes a hard constraint on the KL divergence. To address the limitation of uniform risk, we introduce the \textbf{Signal-Aware Weighted Constraint}. 

Given the space constraints of the column, we formulate the optimization problem as follows:
\begin{equation}
\label{eq:weighted_constraint}
\begin{aligned}
& \max_{\theta} \ \mathbb{E}_t [r_t(\theta) A_t] \\
& \text{s.t.} \quad \mathbb{E}_t \left[ \frac{1}{\rho(A_t, p_g)} D_{KL}(\pi_{\theta_{old}} || \pi_{\theta}) \right] \le \delta
\end{aligned}
\end{equation}
where $\rho(\cdot) \ge 1$ is a scaling factor. A larger $\rho$ reduces the effective cost of the KL divergence, allowing for larger policy updates when signal quality is high.

\subsection{Derivation of the Dynamic Boundary}

Since optimizing the expectation constraint directly is computationally expensive, we convert the global constraint into a local constraint using a second-order approximation.

\paragraph{Taylor Expansion of KL Divergence.}
The KL divergence between the old and new policy can be expressed as the expectation of the negative log-ratio:
\begin{equation}
\begin{aligned}
D_{KL}(\pi_{\theta_{old}} || \pi_{\theta}) &= \mathbb{E}_{x \sim \pi_{\theta_{old}}} \left[ -\log \frac{\pi_{\theta}(x)}{\pi_{\theta_{old}}(x)} \right] \\
&= \mathbb{E}_{x \sim \pi_{\theta_{old}}} [ -\log r_t ]
\end{aligned}
\end{equation}
We perform a second-order Taylor expansion of the function $f(r) = -\log r$ around the point $r = 1$ (since $\pi_{\theta} \approx \pi_{\theta_{old}}$ locally). The derivatives are $f'(1) = -1$ and $f''(1) = 1$. The expansion is:
\begin{equation}
-\log r \approx -(r - 1) + \frac{1}{2}(r - 1)^2
\end{equation}
Substituting this back into the expectation:
\begin{equation}
\begin{aligned}
D_{KL} \approx \ &\mathbb{E}_{x \sim \pi_{\theta_{old}}} [-(r_t - 1)] \\
&+ \mathbb{E}_{x \sim \pi_{\theta_{old}}} \left[ \frac{1}{2}(r_t - 1)^2 \right]
\end{aligned}
\end{equation}
Crucially, the first-order term vanishes because the expectation of the probability ratio is 1:
\begin{equation}
\mathbb{E}[r_t - 1] = \mathbb{E}[r_t] - 1 = \int \pi_{old} \frac{\pi}{\pi_{old}} - 1 = 1 - 1 = 0
\end{equation}
Thus, the KL divergence is dominated by the second-order term:
\begin{equation}
D_{KL}(\pi_{\theta_{old}} || \pi_{\theta}) \approx \frac{1}{2} (r_t - 1)^2
\label{eq:kl_approx}
\end{equation}

\paragraph{Solving for the Boundary.}
We now substitute Eq. \ref{eq:kl_approx} into the local version of the weighted constraint (Eq. \ref{eq:weighted_constraint}). For a specific sample $t$, we require:
\begin{equation}
\frac{1}{\rho_t} \cdot \frac{1}{2} (r_t - 1)^2 \le \delta_{local}
\end{equation}
Rearranging to solve for the maximum allowable deviation $|r_t - 1|$:
\begin{equation}
(r_t - 1)^2 \le 2 \delta_{local} \cdot \rho_t \implies |r_t - 1| \le \sqrt{2 \delta_{local}} \sqrt{\rho_t}
\end{equation}
Defining the baseline clipping threshold as $\epsilon_{base} = \sqrt{2 \delta_{local}}$, we obtain the dynamic threshold $\epsilon_t$:
\begin{equation}
\epsilon_t = \epsilon_{base} \cdot \sqrt{\rho(A_t, p_g)}
\end{equation}
This derivation theoretically justifies why the clipping range should scale with the square root of the signal strength.
\section{Limitations}
The main limitations of this work are the scope of evaluation. First, experiments are limited to medium-scale models, with no validation on larger-scale LLMs (e.g., 10B+ parameters). The computational efficiency, convergence stability, and performance of ETR’s adaptive thresholding when scaled to such sizes remain unexamined. Second, the study focuses solely on structured tasks (e.g., math reasoning) and is not extended to open-ended tasks (e.g., open-domain dialogue, creative writing). The applicability of ETR’s assessment proxies to scenarios with high output diversity and ambiguous evaluation criteria remain unexplored.



\section{Algorithm Pseudocode}
\label{app:algorithm}

Algorithm \ref{alg:etr} outlines the complete training process of GRPO with Elastic Trust Regions. The core innovation lies in lines 8-11, where the clipping threshold is dynamically adjusted based on signal quality.

\begin{algorithm}[t] 
\small 
\caption{Group Relative Policy Optimization with ETR}
\label{alg:etr}
\begin{algorithmic}[1]
\STATE \textbf{Input:} Dataset $\mathcal{D}$, policy $\pi_{\theta}$, ref policy $\pi_{\text{ref}}$, group size $G$.
\STATE \textbf{Hyperparams:} Base clip $\epsilon_{\text{base}}$, weights $(\lambda_1, \lambda_2)$, KL coef $\beta$.

\FOR{each training step}
    \STATE Sample query $q \sim \mathcal{D}$.
    \STATE Generate outputs $O = \{o_1, \dots, o_G\}$ via $\pi_{\theta}(\cdot|q)$.
    \STATE Compute rewards $\mathbf{r} = \{r_1, \dots, r_G\}$.
    
    \STATE \textit{// 1. Compute Group Statistics (Macro)}
    \STATE $p_g \leftarrow \frac{1}{G} \sum_{i=1}^G \mathbb{I}(r_i > 0)$.
    \STATE $\mathcal{G}_{\text{macro}} \leftarrow \lambda_2 \cdot 4 p_g (1 - p_g)$.
    
    \STATE \textit{// 2. Compute Advantages (Standard GRPO)}
    \STATE $A_i \leftarrow (r_i - \text{mean}(\mathbf{r})) / (\text{std}(\mathbf{r}) + \xi)$ for all $i$.
    
    \STATE \textit{// 3. Compute Dynamic Thresholds (Micro)}
    \STATE Initialize effective clips $\mathcal{E} \leftarrow \emptyset$.
    \FOR{$i = 1$ to $G$}
        \STATE $\mathcal{S}_{\text{micro}} \leftarrow \lambda_1 \cdot \tanh(A_i)$.
        \STATE $\epsilon_i \leftarrow \epsilon_{\text{base}} + \mathcal{S}_{\text{micro}} + \mathcal{G}_{\text{macro}}$.
        \STATE $\mathcal{E} \leftarrow \mathcal{E} \cup \{\epsilon_i\}$.
    \ENDFOR
    
    \STATE \textit{// 4. Policy Update}
    \STATE Compute loss $\mathcal{J}_{\text{ETR}}$ using dynamic bounds $\mathcal{E}$ (Eq.~\ref{eq:etr_epsilon}).
    \STATE Update $\pi_{\theta}$ via gradient descent $\nabla_{\theta} \mathcal{J}_{\text{ETR}}$.
\ENDFOR
\end{algorithmic}
\end{algorithm}
\section{Implementation Details}
\label{app:implementation}

\subsection{Experimental Setup}
Our experiments are conducted using the \texttt{verl} framework~\citep{sheng2025hybridflow} on a computational node equipped with $8 \times$ NVIDIA H20 GPUs. We utilize the DAPO-Math-17k dataset for training across all models. The prompt templates follow the standard chat formats corresponding to each base model (e.g., Qwen-Chat or Llama-Instruct).

\subsection{Hyperparameters}
Table \ref{tab:main_hyperparams} lists the detailed hyperparameters used in our experiments. To ensure fair comparison, we maintain consistent settings between GRPO and ETR, differing only in the clipping mechanism.

For the model-specific configurations, we adjust the \texttt{max\_response\_length} to accommodate the capacity of different architectures:
\begin{itemize}
    \item Qwen3-8B-Base and Llama-3.1-8B-Instruct: Set to 8192 tokens.
    \item Qwen2.5-Math-7B-Base: Set to 3584 tokens due to its embedding positions limits.
\end{itemize}

For our proposed ETR method, we set the elasticity coefficients $\lambda_1 = 0.1$ (micro-level) and $\lambda_2 = 0.1$ (macro-level) across all experiments without further tuning.

\begin{table}[htbp]
\centering
\small
\caption{Hyperparameters for training and evaluation. ETR introduces $\lambda_1$ and $\lambda_2$, while other settings remain identical to the GRPO baseline.}
\label{tab:main_hyperparams}
\begin{tabular}{l l}
\toprule
\textbf{Hyperparameter} & \textbf{Value} \\
\midrule
\multicolumn{2}{l}{\textit{General Training Config}} \\
Optimizer & AdamW \\
Learning Rate & 1e-6 (Constant) \\
Global Batch Size & 64 \\
Rollouts per Prompt ($G$) & 8 \\
Gradient Clipping & 1.0 \\
KL Coefficient ($\beta$) & 0.001 \\
Reward Function & $\pm 1$ (Binary) \\
\midrule
\multicolumn{2}{l}{\textit{Generation Config}} \\
Training Temperature & 1.0 \\
Training Top-p & 1.0 \\
Evaluation Temperature & 1.0 \\
Max Response Length & \{8192, 3584\} \\
\midrule
\multicolumn{2}{l}{\textit{ETR Specifics}} \\
Base Clip ($\epsilon_{base}$) & 0.2 \\
Micro-Elasticity ($\lambda_1$) & \textbf{0.1} \\
Macro-Elasticity ($\lambda_2$) & \textbf{0.1} \\
\bottomrule
\end{tabular}
\end{table}
\section{Analysis}
\label{sec:analysis}

In this section, we analyze the training dynamics to understand the source of ETR's performance gains. We focus on the clipping behavior and response length evolution, while referring back to the accuracy curves presented in Figure \ref{fig:acc_curves}.

\begin{figure*}[t]
    \centering
    \begin{minipage}{0.48\linewidth}
        \centering
        \includegraphics[width=\linewidth]{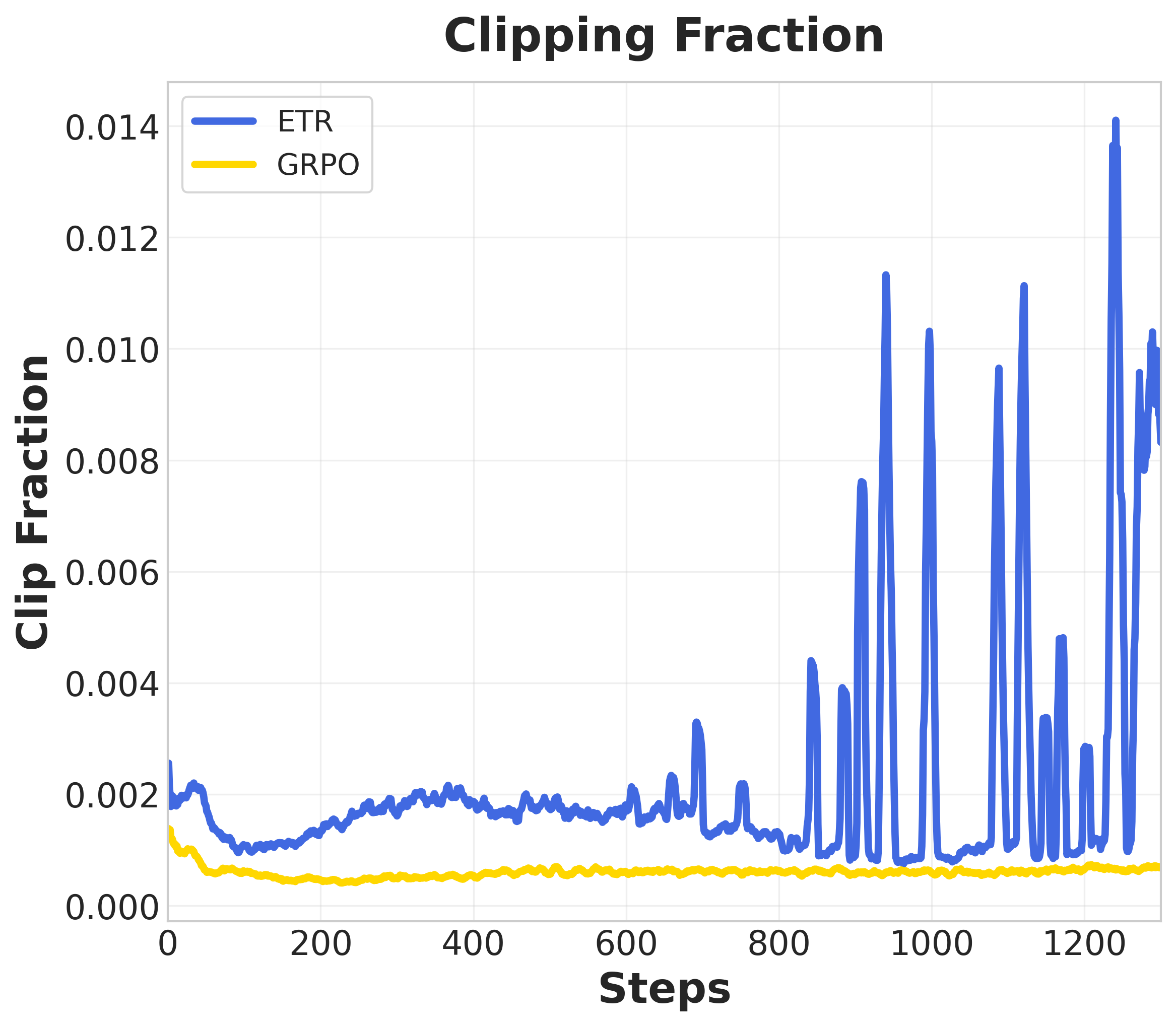} 
        \caption{\textbf{Clipping Fraction.} ETR triggers clipping significantly more often than GRPO. This reflects the \textit{tightening} of constraints on negative samples, effectively filtering out diffusion noise from incorrect paths.}
        \label{fig:clip_fraction}
    \end{minipage}
    \hfill
    \begin{minipage}{0.48\linewidth}
        \centering
        \includegraphics[width=\linewidth]{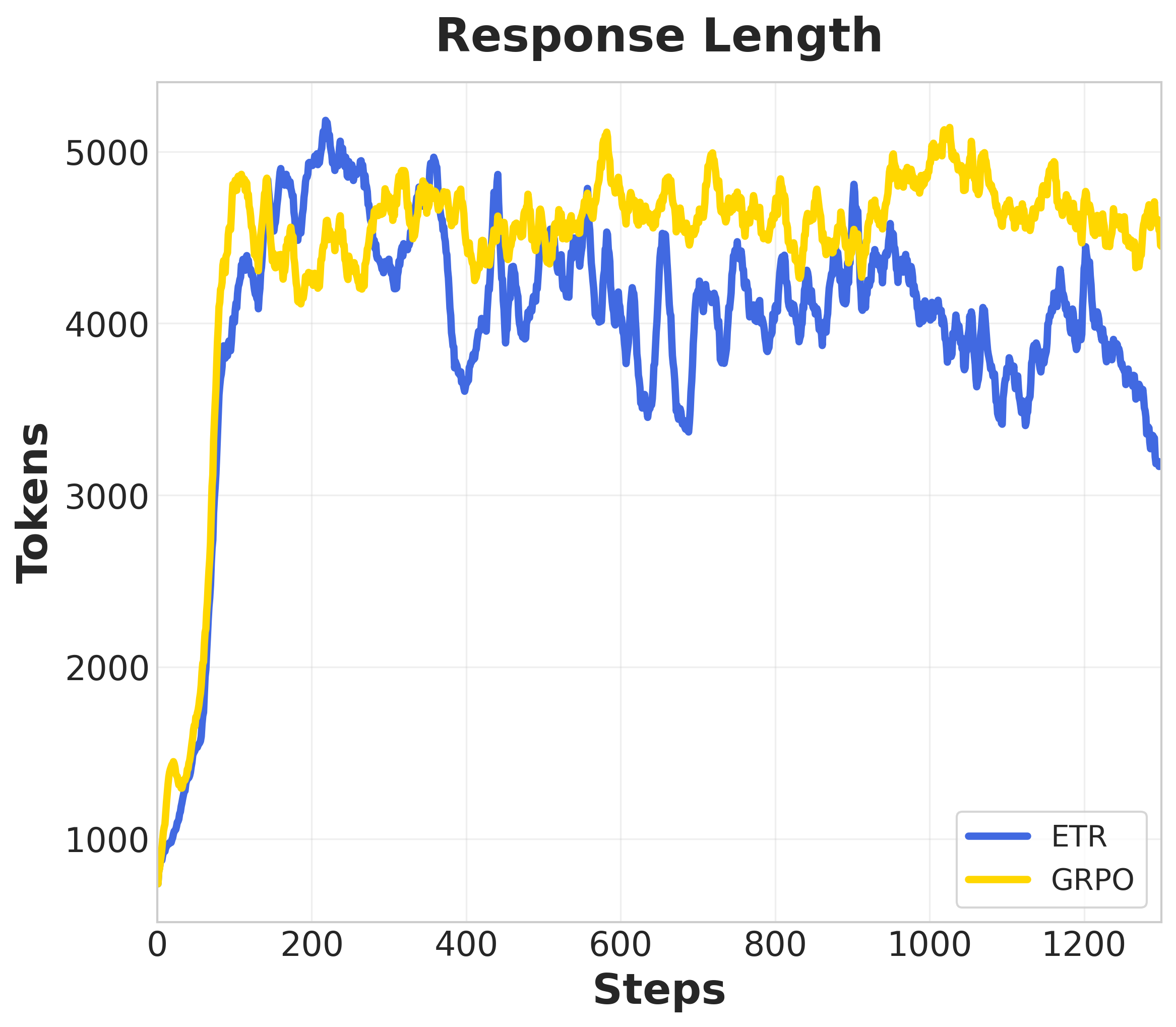} 
        \caption{\textbf{Response Length.} ETR (Blue) effectively curbs the "length hacking" phenomenon observed in GRPO (Yellow), promoting more efficient reasoning chains.}
        \label{fig:response_length}
    \end{minipage}
\end{figure*}

\subsection{Asymmetric Noise Suppression}
Figure \ref{fig:clip_fraction} illustrates the fraction of samples where the policy update is clipped. A counter-intuitive observation is that ETR triggers clipping much more frequently (spikes $>1.0\%$) than the static GRPO baseline (stable at $\approx 0.1\%$), despite ETR's ability to expand the trust region.

This phenomenon validates the \textbf{asymmetric design} of our dynamic boundary. For negative samples ($A_t < 0$), the threshold tightens ($\epsilon_t < \epsilon_{base}$). Since incorrect reasoning paths typically constitute the majority of generated data in difficult reasoning tasks, these negative samples frequently hit the tightened boundary. 
Therefore, the high clipping fraction indicates that ETR is actively \textbf{suppressing noise}. By strictly limiting the gradient updates from erroneous paths, ETR prevents the model from "unlearning" useful logic due to low-quality negative signals, while implicitly reserving the trust region budget for the sparser, positive signals where the boundary is expanded.

\subsection{Regularization against Reward Hacking}
Figure \ref{fig:response_length} reveals a common failure mode in RLVR: \textit{Reward Hacking via Length}. The GRPO baseline (Yellow) rapidly drifts towards generating excessively long responses ($>5000$ tokens), likely attempting to maximize the probability of hitting a correct answer through verbose exploration. This bloating increases training latency without proportional accuracy gains.

In contrast, ETR (Blue) acts as a geometric regularizer. When the model generates long, incorrect chains, the advantage is negative, causing the trust region to shrink. This penalizes the model heavily if it drifts too far from the reference policy on wrong paths. As a result, ETR stabilizes the response length at a more efficient level ($\approx 3500$ tokens), encouraging concise and precise reasoning.

\subsection{Prevention of Policy Collapse}
As shown previously in \textbf{Figure \ref{fig:acc_curves}}, standard GRPO suffers from a "performance collapse" in the later stages of training, where the \textit{Best@32} accuracy degrades significantly. This collapse is often linked to the unchecked accumulation of policy shifts induced by noisy updates.

By implementing the asymmetric constraints described above—tightening on errors and relaxing on successes—ETR maintains policy stability. The dynamic boundary ensures that the policy does not drift excessively on ambiguous or incorrect samples. Consequently, ETR sustains a robust upward trajectory in accuracy throughout the training process, avoiding the degradation seen in the baseline.

\end{document}